\documentclass[10pt,twocolumn,letterpaper]{article}

\usepackage{iccv}
\usepackage{times}
\usepackage{epsfig}
\usepackage{graphicx}
\usepackage{amsmath}
\usepackage{amssymb}

\usepackage{multirow}
\usepackage{makecell}
\usepackage{amsmath,amssymb}
\usepackage{color}
\usepackage{subcaption}

\usepackage[breaklinks=true,bookmarks=false]{hyperref}

\iccvfinalcopy 

\def\iccvPaperID{****} 

\ificcvfinal\fi

\begin{document}

\title{Normalization Matters in Weakly Supervised Object Localization}

\author{Jeesoo Kim\textsuperscript{\rm 1} \quad Junsuk Choe\textsuperscript{\rm 2, 3} \quad Sangdoo Yun\textsuperscript{\rm 2} \quad Nojun Kwak\textsuperscript{\rm 1} \\
\\
\small{\textsuperscript{\rm 1} Department of Intelligence and Information, Seoul National University
\quad
\textsuperscript{\rm 2} NAVER AI Lab}}

\maketitle
\ificcvfinal\thispagestyle{empty}\fi

\begin{abstract}
   Weakly-supervised object localization (WSOL) enables finding an object using a dataset without any localization information.
   By simply training a classification model using only image-level annotations, the feature map of the model can be utilized as a score map for localization.
   In spite of many WSOL methods proposing novel strategies, there has not been any de facto standard about how to normalize the class activation map (CAM).
   Consequently, many WSOL methods have failed to fully exploit their own capacity because of the misuse of a normalization method.
   In this paper, we review many existing normalization methods and point out that they should be used according to the property of the given dataset.
   Additionally, we propose a new normalization method which substantially enhances the performance of any CAM-based WSOL methods.
   Using the proposed normalization method, we provide a comprehensive evaluation over three datasets (CUB, ImageNet and OpenImages) on three different architectures and observe significant performance gains over the conventional min-max normalization method in all the evaluated cases (See Fig.~\ref{fig:intro}).
\end{abstract}

\section{Introduction}
Given nothing but the class information of an object, \textit{weakly-supervised object localization} (WSOL) allows a convolutional neural network (CNN) to localize the object in a scene.
Although many fully-supervised object detectors guarantee considerable performance in locating objects in an image, localization techniques in the absence of bounding box annotations are still in need.

WSOL using neural networks has been initially introduced by the class activation map (CAM)~\cite{zhou2016learning} approach.
Training a convolutional neural network (CNN) model with a classification problem enables the model to generate an activation map from the last layer of it.
After that, simply cutting out the activation map with a proper threshold enables the localization of an object.
In spite of the plausible axiom that features contributing better to a specific class are likely to represent the location of the object, the problem still exists that the discriminative parts of an object hoax the activation of the model to make an inaccurate localization of the target object.
Many methods have been proposed to overcome this problem \cite{singh2017hide,zhang2018adversarial,zhang2018self,choe2019attention,yun2019cutmix} and persuasive evidences of performance improvement have been demonstrated qualitatively and quantitatively.

\begin{figure}[t]
\begin{center}
   \includegraphics[width=\linewidth]{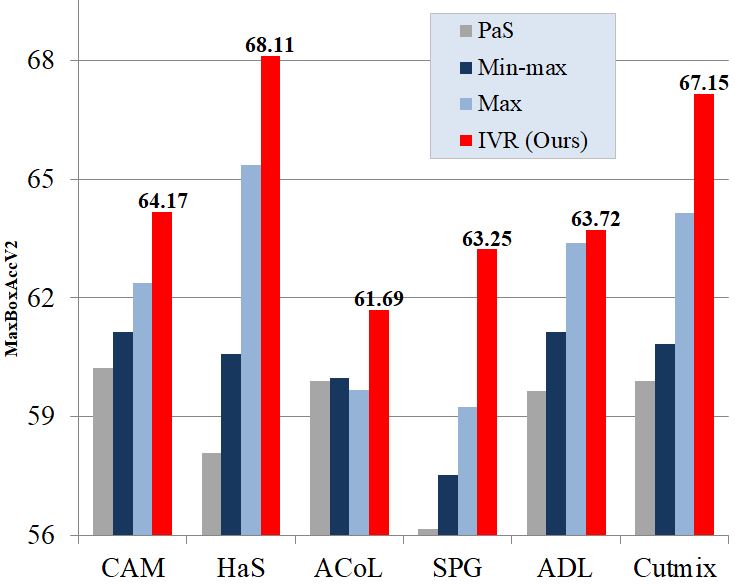}
\end{center}
   \caption{\small{Comparison of several WSOL methods with different kinds of normalization methods for a class activation map. The accuracy has been evaluated under the evaluation metric suggested in \cite{choe2020evaluating} with CUB-200-2011 dataset. All scores in this figure are the average scores of ResNet50, VGG16, and InceptionV3. In all WSOL methods, the performance using our normalization method, IVR, is the best.}}
\label{fig:intro}
\end{figure}

However, prior works have been evaluated under different conditions and their hyperparameters have been chosen empirically.
Usually, the feature from the last layer of the model is post-processed to be used as a class activation map.
The normalization scheme used by every method differs each other and makes the comparison unfair.
The work of \cite{choe2020evaluating} has proposed a new evaluation protocol and offered a thorough comparison of six previous WSOL methods.
In their experiments, the min-max normalization has been applied to every method for the fair comparison and the best hyperparameters have been found using a random search \cite{bergstra2012random}.
According to it, all methods have shown almost no improvement compared to the original CAM \cite{zhou2016learning}.

In the work of \cite{bae2020rethinking}, the authors suggest a thresholding strategy that excludes exceptionally high activation values in each image.
In the sense that the valid value range in the activation map changes, this can be compared with other normalization methods.
With this new method, some methods such as CAM \cite{zhou2016learning} and HaS \cite{singh2017hide} have shown performance improvements in several datasets.

In this paper, we investigate the problem that can occur when using the min-max normalization.
Even though the min-max normalization is the most popular scheme in recent WSOL works, we verify that the min-max scheme can deteriorate the performance of most WSOL methods.
We revisit max normalization which has been prevalent for a long time and point out that it can resolve the problem above.
Also, although other methods including \textit{percentile as a standard for thresholding} (PaS)~\cite{bae2020rethinking} have shown better results than the min-max normalization, they still suffer from problems which will be described in this paper.
To resolve these problems, we propose a new normalization method \textit{inferior value removal} (IVR).
Through extensive experimental results, IVR has been shown to improve the localization performances of almost all WSOL methods.
Enabling better exploitation of each WSOL method, a comprehensive re-evaluation and ordering of several WSOL methods have been conducted.
The contribution of this paper is as follows:

\begin{itemize}
    \item We provide a thorough investigation about problems of commonly used normalization methods in WSOL. The problem originating from using min-max normalization is explained qualitatively and quantitatively.
    \item We propose a new normalization method which can better exploit the performance of many WSOL methods. It can be used in any kind of WSOL methods which use a class activation map.
    \item A comprehensive evaluation with various kinds of normalization methods in three different datasets and three different architectures has been made. We provide a renewed benchmark of six WSOL methods.
\end{itemize}

\section{Related works}
Locating objects in images is one of the most important and frequently studied tasks in the field of computer vision.
Depending on the type of supervision given to the model, semantic segmentation \cite{zhu2019improving,zhao2017pyramid,chen2018encoder,lin2019zigzagnet,fu2019adaptive} and object detection \cite{ren2015faster,liu2016ssd,redmon2016you,lin2017focal,bochkovskiy2020yolov4,zhang2020bridging,tan2020efficientdet} can be powerful localization methods.

There has been a constant demand on a localization method relying only on the image-level annotation.
Class activation mapping (CAM) \cite{zhou2016learning} is the first approach using the activated convolutional feature as a score map to locate an object in an image.
As classification is not designed for a localization task, CAM often fails to successfully capture the whole object extent.
However, thanks to its availability, many researchers have evolved CAM using various training strategies.
Hide-and-seek (HaS) \cite{singh2017hide} makes a grid in an image and randomly erases multiple patches.
The model struggles to make a correct decision with the corrupted image and this induce the feature of the model to be activated in the location of the target object.
Adversarial complementary learning (ACoL) \cite{zhang2018adversarial} uses two branches which adversarially get rid of the highlighted activated region from each other.
This approach is different from HaS in that the feature is erased instead of the image itself.
Self-produced guidance (SPG) \cite{zhang2018self} generates activation masks in each layer and uses them as a pseudo supervision for the preceding layer.
Attention-based dropout layer (ADL) \cite{choe2019attention} adds attention modules to the model and adversarially drops highly activated regions.
CutMix \cite{yun2019cutmix} has originally been designed to enhance the robustness of any CNN model.
Patches from training images are cut and pasted to one another and this helps the model to capture less discriminative parts of the target object.
Aside from CAM-based WSOL methods mentioned above, various gradient-based WSOL methods \cite{selvaraju2017grad,chattopadhay2018grad,simonyan2013deep,samek2016evaluating} have been proposed as well.
However, they are often heavy to be used practically and we do not consider these methods in this paper.

Meanwhile, a comprehensive and fair evaluation of CAM-based methods mentioned above has been made in \cite{choe2020evaluating}.
More reasonable evaluation metrics have been proposed and all the considered WSOL methods have been re-evaluated with a thorough hyperparameter search and a unified min-max normalization method.
As a result, it has been claimed that all WSOL methods after the emergence of CAM have turned out to be not significantly different in performance from CAM. 
To make a better use of the class activation map, a percentile based threshold modification method has been proposed \cite{bae2020rethinking}.
This method assumes that large values must be treated as outliers and excluded from the class activation map.
With the help of other techniques, several existing methods have succeeded to improve the performance.
However, empirical choice of threshold violates the evaluation protocol in \cite{choe2020evaluating} that every WSOL method has its own optimal threshold.
In this paper, we review the problems which most CAM-based WSOL methods experience and reset the benchmark for the future research.

\section{Preliminary}
WSOL is a task of finding an object belonging to a specific class in an image.
While training a CNN with a classification problem, intermediate layers inside the model generate activation patterns to provide a correct output.
After a given image $X \in \mathbb{R}^{W \times H \times 3}$ is processed by a CNN $f$,  its generated feature map is usually average-pooled and handed to a classifier.
A weight vector $\mathbf{w}^{c}_i$ from the classifier is multiplied to the average-pooled feature and then the logit of the class $c$ becomes available.
On the other hand, by multiplying back this weight vector $\mathbf{w}_i^c$ to the activated feature map $f(X) \in \mathbb{R}^{W'\times H' \times K}$, we can infer the location of an object belonging to class $c$ within an image.
After the multiplication, averaging along the channel dimension produces $\mathbf{F} \in \mathbb{R}^{W' \times H'}$, which we call as \textit{class activation map} in this paper: 
\begin{equation}
    \mathbf{F}_c=\frac{1}{K}\sum_{i=1}^K\mathbf{w}^{c}_{i}\cdot f_{i}(X)
    \label{eq:wsol}
\end{equation}
\begin{equation}
    \mathbf{F}_c'=H(\mathbf{F}_c).
    \label{eq:norm}
\end{equation}
Here, $K$ denote the number of channels. From now on, we omit the subscript $c$ denoting the class index for brevity.

In all WSOL methods, values in $\mathbf{F}$ typically lie in a certain range whose minimum value is a negative value in most cases and maximum value may be either a very small or large positive value.
To apply a consistent level of threshold to all images, $\mathbf{F}$ is normalized by a normalization function $H$ into the range of $[0, 1]$ producing a score map $\mathbf{F'}$.
Not many of researchers have looked upon the importance of feature normalization.
We discuss the effect of many different normalization functions $H$ in this paper and suggest strategies for a better evaluation of WSOL.

\section{Normalizing the class activation map}
The context of images in a dataset or the portion of an object inside the image can induce a substantially different range of values in the score map $\mathbf{F'}$.
Many normalization methods have been suggested but how the normalization can affect the performance has never been dealt with.
Here in this section, we explain all the existing normalization methods up to our knowledge including the one we propose.

\begin{figure}[t]
\begin{center}
   \includegraphics[width=\linewidth]{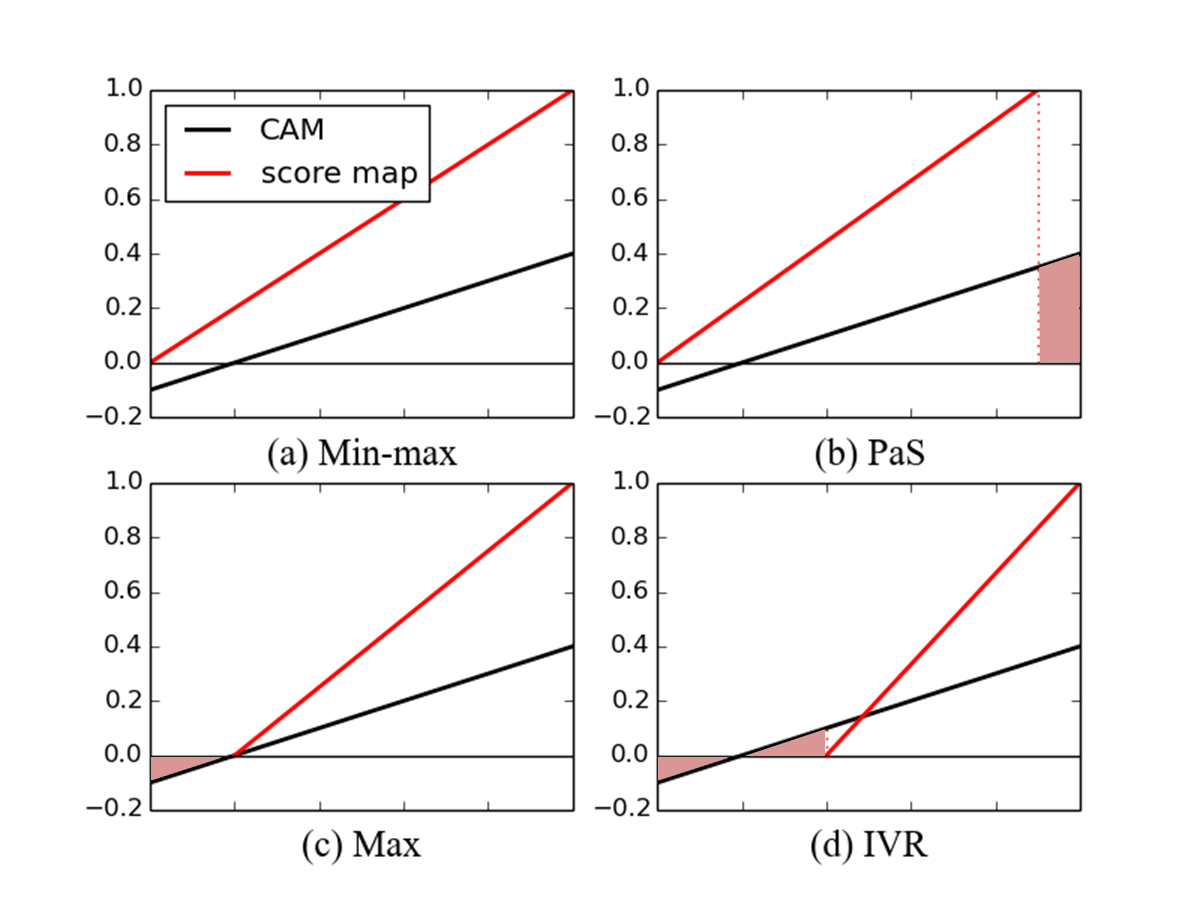}
\end{center}
   \caption{Illustration of how each normalization method aligns values in the class activation map. In this case, all methods share the same class activation map $\mathbf{F}$ whose range is assumed to be [-0.1, 0,4].
   Different methods map $\mathbf{F}$ into different score maps $\mathbf{F'}$ all ranging from 0 to 1.
   Consequently, we can assume that the region colored in red is excluded from the final score map.}
\label{fig:norm_methods}
\end{figure}

\begin{figure*}[t]
     \centering
     \begin{subfigure}[b]{0.43\textwidth}
         \centering
         \includegraphics[width=\textwidth]{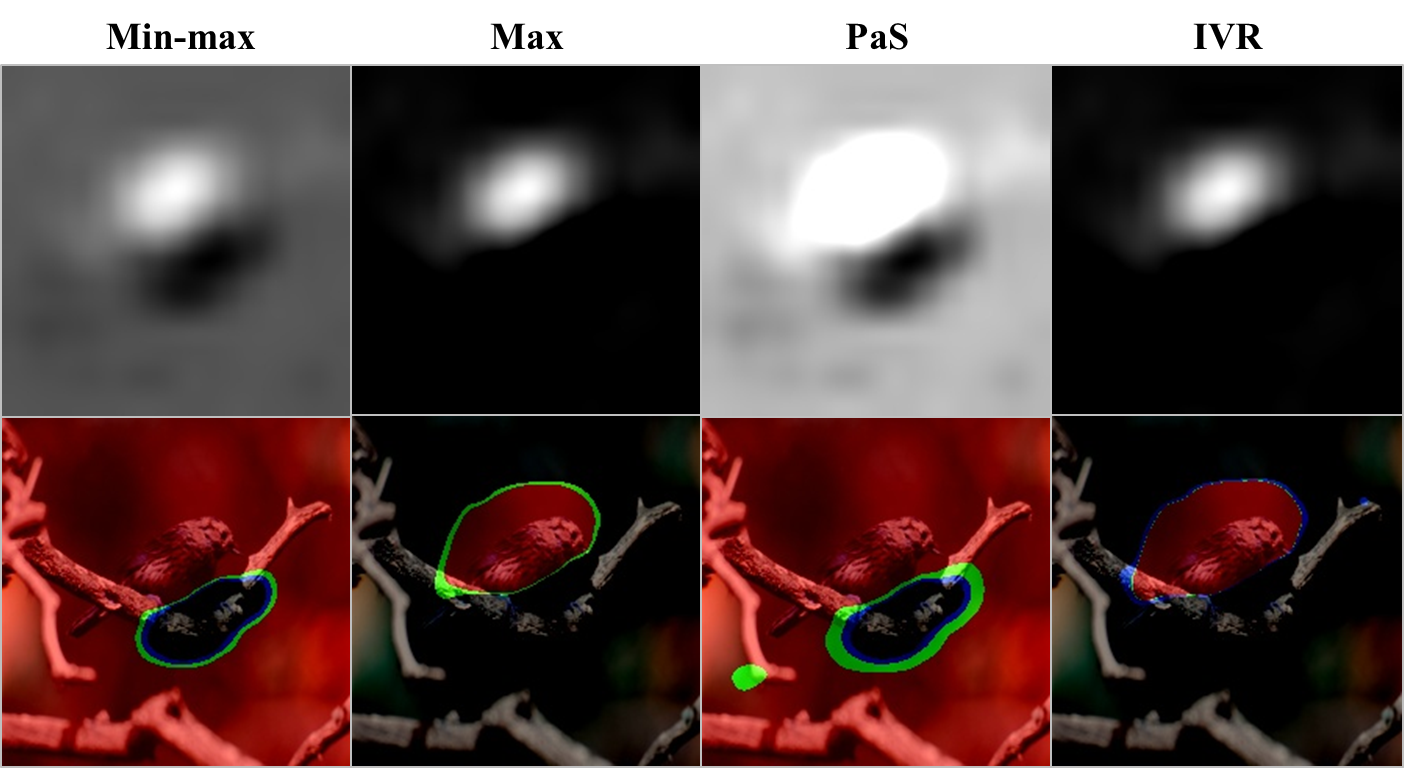}
          \caption{CAM}
     \end{subfigure}
     \begin{subfigure}[b]{0.43\textwidth}
         \centering
         \includegraphics[width=\textwidth]{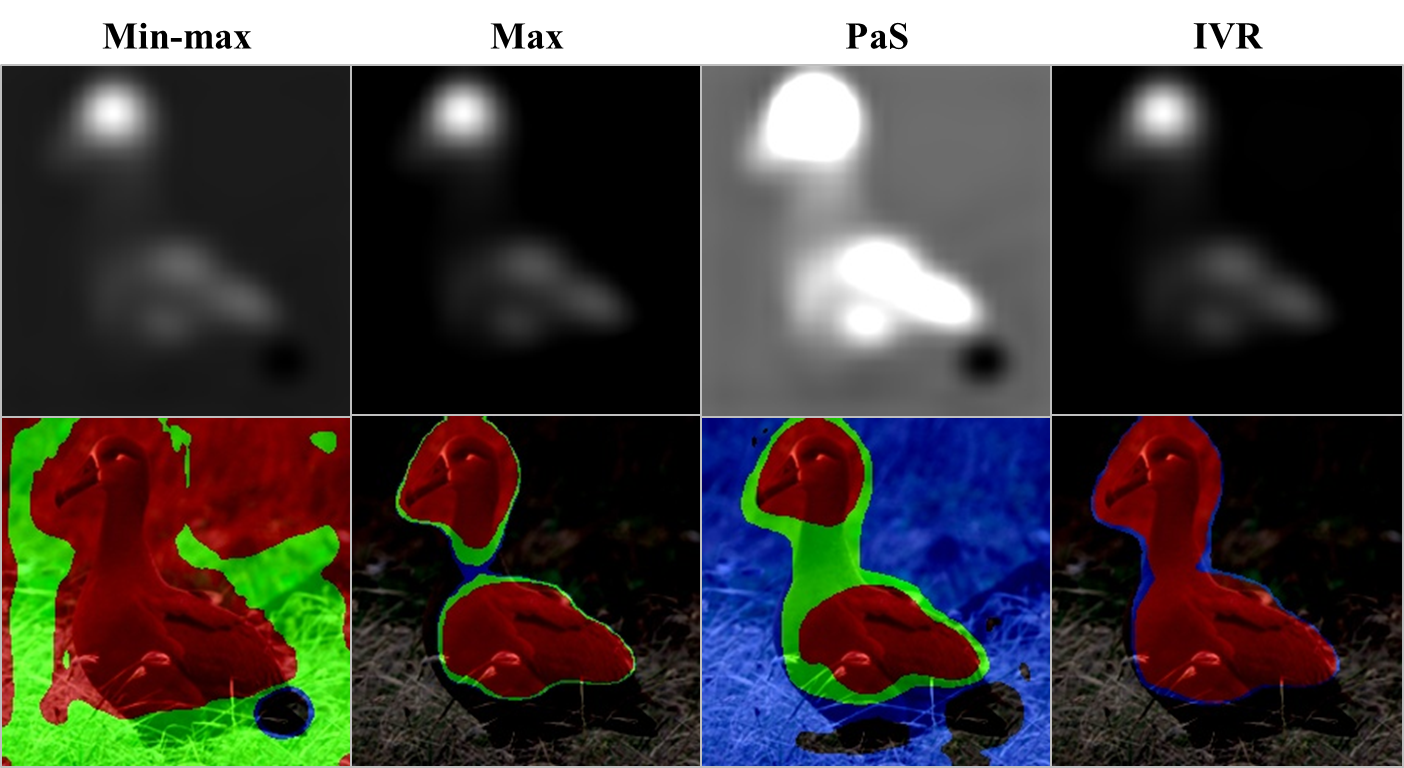}
          \caption{HaS}
     \end{subfigure}
     \begin{subfigure}[b]{0.43\textwidth}
         \centering
         \includegraphics[width=\textwidth]{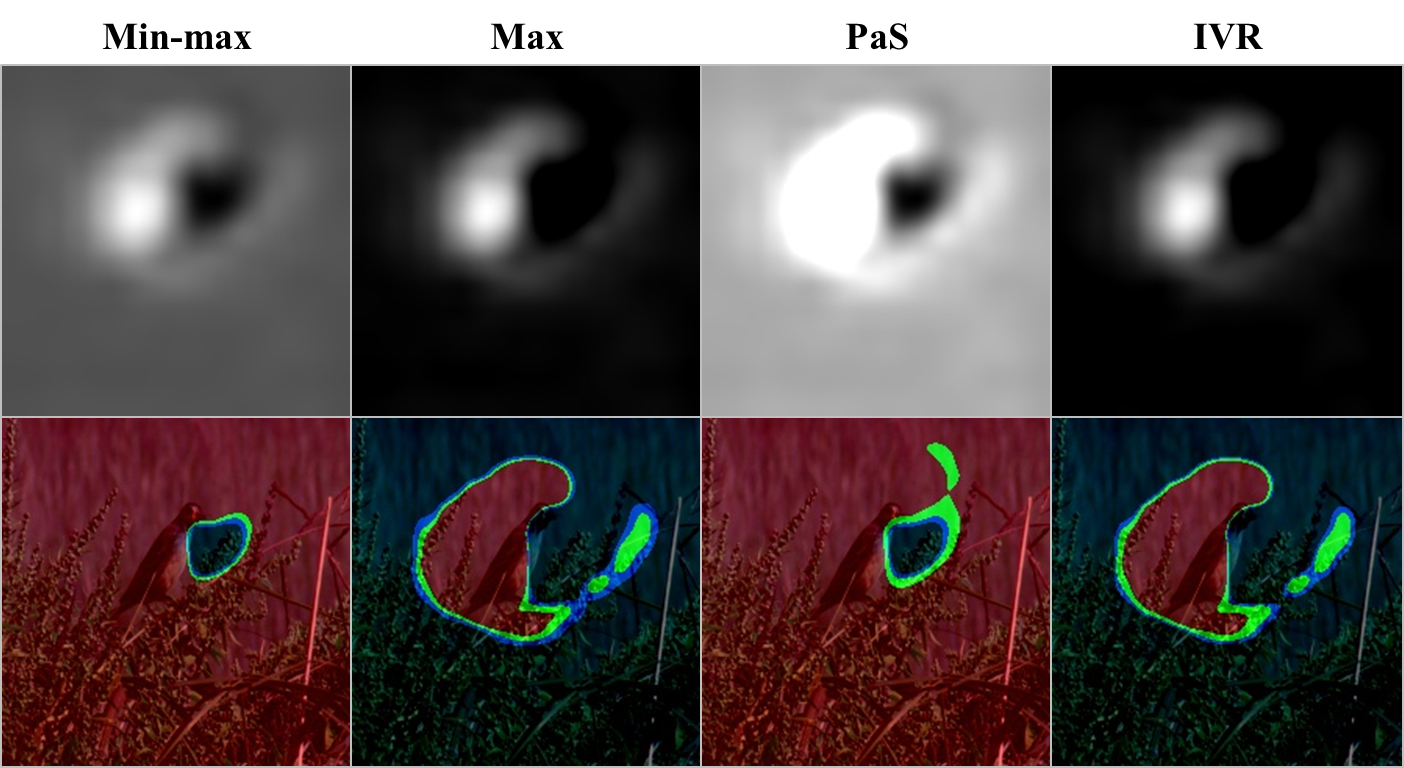}
          \caption{ADL}
     \end{subfigure}
     \begin{subfigure}[b]{0.43\textwidth}
         \centering
         \includegraphics[width=\textwidth]{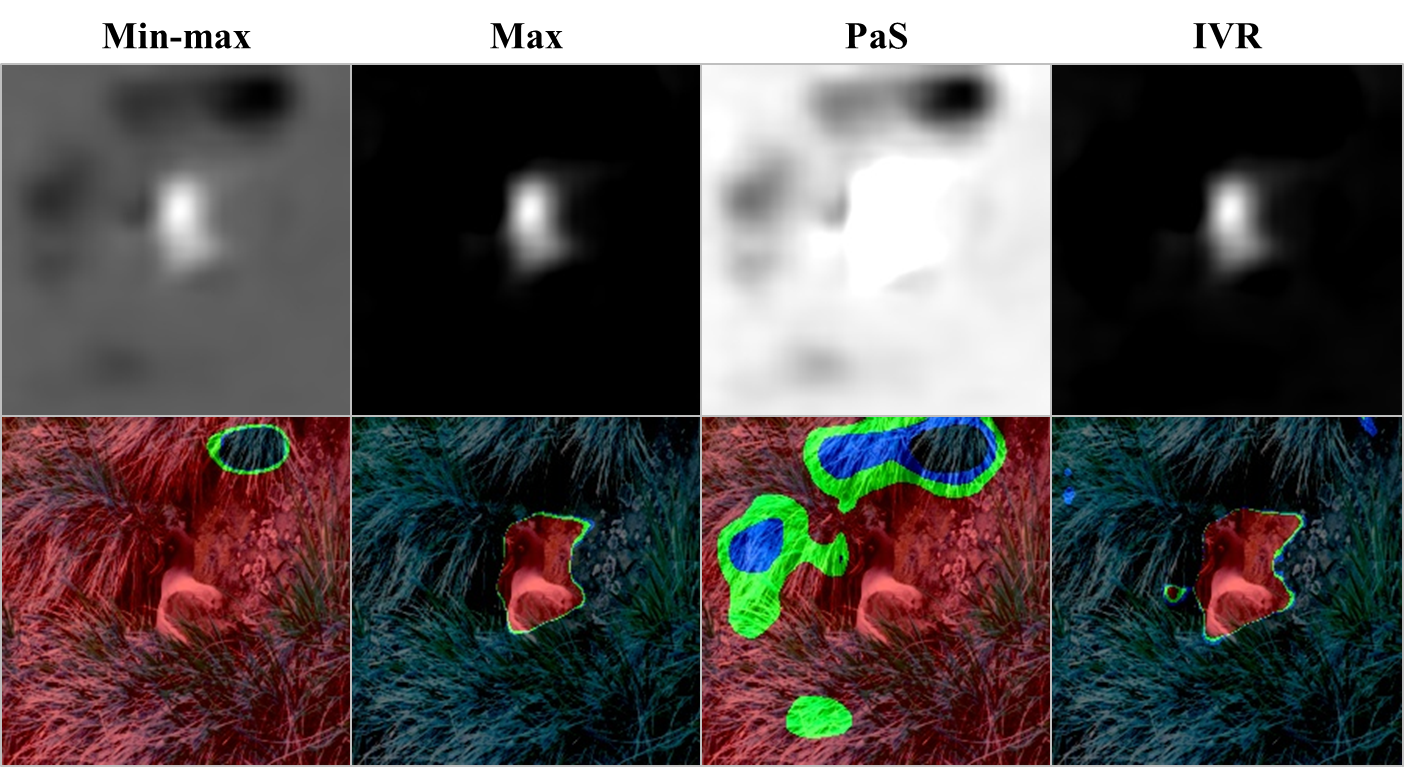}
          \caption{CutMix}
     \end{subfigure}
        \caption{Visualization of score maps and activated areas by the optimal thresholds in CAM, HaS, ADL, and CutMix on CUB dataset. Red, green and blue area are regions extracted by the optimal thresholds of IoU 0.3, 0.5 and 0.7.}
        \label{fig:camfeature}
\end{figure*}

\subsection{Min-max normalization}
Min-max normalization is the most frequently used normalization method for WSOL which has been used in ACoL \cite{zhang2018adversarial}, SPG \cite{zhang2018self}, ADL \cite{choe2019attention} and CutMix \cite{yun2019cutmix}.
After the minimum value is subtracted from the class activation map, it is divided by its maximum value.
This can be expressed as follows:
\begin{equation}
    \mathbf{F'}=\frac{\mathbf{F}-\min(\mathbf{F})}{\max(\mathbf{F})-\min(\mathbf{F})}.
    \label{eq:minmax}
\end{equation}

By aligning the minimum value to zero and the maximum to one, all values in the score maps fall within the score range.
This is described visually in (a) of Fig.~\ref{fig:norm_methods}.
To provide a fair comparison under a consistent normalization method, the work of \cite{choe2020evaluating} suggests a well designed evaluation metric and compares every methods under min-max normalization.
The conclusion of \cite{choe2020evaluating} is that the WSOL methods that have emerged after CAM actually do not show a significant performance enhancement.
However, we observed that using min-max normalization experiences a problem that has not been considered in the work of \cite{choe2020evaluating}.
Outlier values which are too large or small may distort other values during normalization.
Within a dataset, class activation maps of some images (but not rare) often include exceptionally smaller minimum values and this often significantly drops the localization performance on those images.
For most images which have an ordinary level of minimum value in the class activation map $\mathbf{F}$, this minimum value does not affect the overall score map $\mathbf{F'}$.
However, when it comes to images that have an exceptionally smaller minimum value, it raises the activation level of the whole image and almost the entire image region is localized as the object in question (See Fig. \ref{fig:camfeature}, especially, CAM, HaS and CutMix).

\subsection{Max normalization}
Max normalization has been first used in CAM \cite{zhou2016learning} and then in HAS \cite{singh2017hide}.
Unlike min-max normalization, max normalization divides feature values by their maximum value and constrains the class activation map to be lower than one.
This is described visually in (c) of Fig.~\ref{fig:norm_methods}.
This can be expressed as follows:
\begin{equation}
    \mathbf{F'}=\frac{\mathbf{F}}{\max(\mathbf{F})}
    \label{eq:max}
\end{equation}

Negative values may exist 
after the normalization.
Allowing values below zero, this method linearly transforms $\mathbf{F}$ into $\mathbf{F'}$. 
As typically the optimal activating threshold $\tau^*$ of a WSOL method is expected to be positive, max normalization can be considered simply as a method that ignores negative values.
However, the value range of all images differ considerably and there is possibility that negative values may contribute to the quality of score maps. 

\subsection{Percentile as a Standard for Thresholding (PaS)}
Percentile as a Standard for Thresholding (PaS) has been proposed in \cite{bae2020rethinking}.
The authors say that exceptionally large values in the activation map make other relatively small values to be ignored.
These large values must be excluded from the score map by using a percentile instead of their maximum value.
This is described visually in (b) of Fig.~\ref{fig:norm_methods}.
According to the code\footnote{https://github.com/won-bae/rethinkingCAM} available by the author, PaS uses min-max normalization except the maximum value is substituted with 90 percentile of all score map values.
Therefore, it can be seen as a variety of the min-max normalization.
This can be expressed as follows:
\begin{equation}
    \mathbf{F'}=\frac{\mathbf{F}-\min(\mathbf{F})}{\textrm{Pct}_{p}(\mathbf{F}-\min(\mathbf{F}))}.
    \label{eq:pas}
\end{equation}

$\textrm{Pct}_{p}(\cdot)$ denotes a function which finds $p$ percentile from the given values.
In the original paper, the optimal threshold $\tau^*$ is fixed to a specific value for every dataset and $p$ is set to be 90 heuristically.
However, the evaluation metric suggested in the work of \cite{choe2020evaluating} validates all threshold values and picks the best threshold since WSOL methods are quite vulnerable to the un-optimized threshold.
We re-evaluate PaS based on this metric and report the performance of this method.
Also, we further apply PaS in all other WSOL methods and datasets mentioned in other papers \cite{bae2020rethinking,choe2020evaluating}.

\begin{figure*}[t]
\begin{center}
   \includegraphics[width=\linewidth]{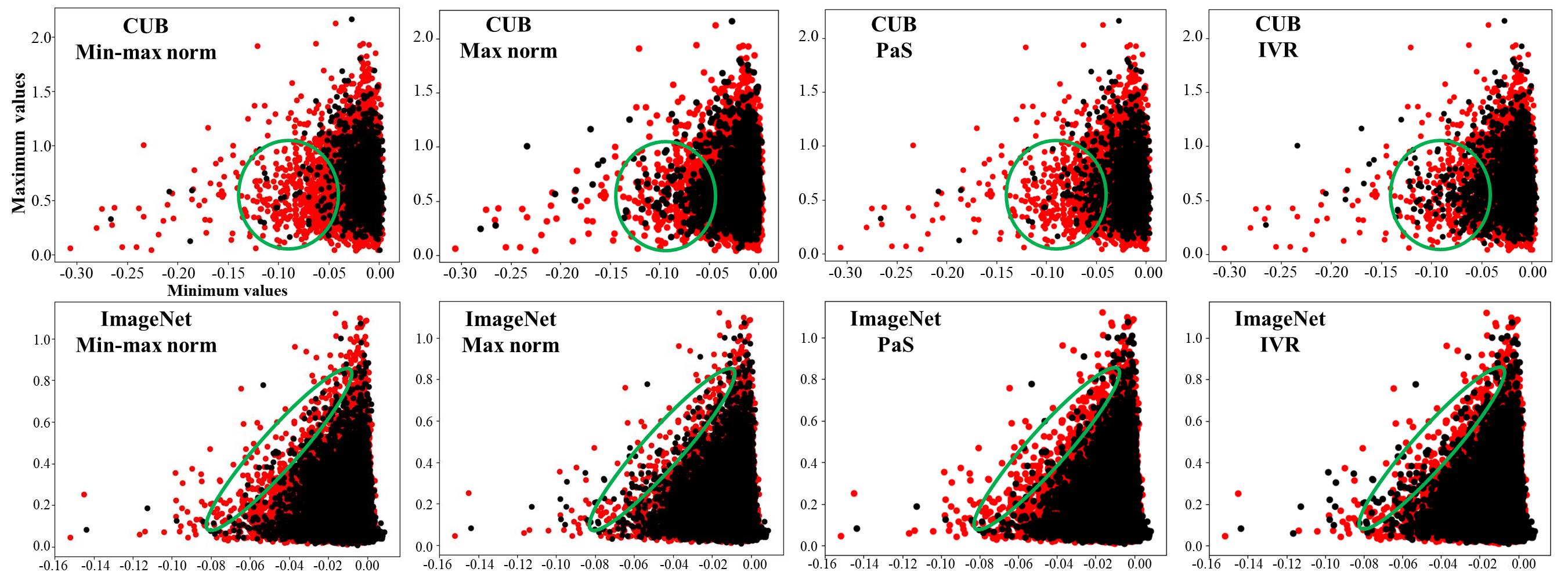}
\end{center}
   \caption{Distribution of minimum and maximum values from class activation map $F$ of all test images. Each point represents a single image whose values from horizontal and vertical axes are minimum value and maximum value respectively. Black dots correspond to correctly localized images while red dots are not. 
   Note that IVR shows overall higher density of positive samples (black dots) than other methods as highlighted with green circles. 
   }
\label{fig:minmaxdist}
\end{figure*}

\subsection{Inferior Value Removal (IVR)}
In this paper, we propose a new normalization method, Inferior Value Removal (IVR), with a thorough investigation.
The basis of IVR is that extremely small values in the class activation map unnecessarily raise the overall values in the score map and this disturbs the consistent thresholding among images.
Unlike PaS, IVR is a normalization methods which restricts the minimum value of the activation map.
Before dividing all values in $F$ by their maximum value, a percentile value from its minimum value is subtracted.
This is described visually in Fig.~\ref{fig:norm_methods} (d).
In other words, IVR uses max normalization except the minimum value is substituted with a certain percentile value of all activation map values.
This can be expressed as follows:
\begin{equation}
    \mathbf{F'}=\frac{\mathbf{F}-\textrm{Pct}_{p}(\mathbf{F})}{\max(\mathbf{F}-\textrm{Pct}_{p}(\mathbf{F}))}.
    \label{eq:IVR}
\end{equation}

Values up to $p$-percentile are excluded from the activation map and the remaining values are re-arranged by their maximum value.
The percentile parameter $p$ is decided from the validation set and the performance tendency has been verified to be the same with the test set.
By using IVR, we can use the original activation map which is relatively unaffected from the value shift by its minimum value while the range is calibrated according to the given values.

\section{Experiments}
\subsection{Settings}
\noindent
\textbf{Dataset:}
For the evaluation of WSOL, Caltech-UCSD Birds-200-2011 (CUB) \cite{welinder2010caltech}, ImageNet \cite{russakovsky2015imagenet} and OpenImages~\cite{OpenImages} are most frequently used.
In the work of \cite{choe2020evaluating}, both datasets are separated into three different divisions: \texttt{train-weaksup, train-fullsup} and \texttt{test}.
We follow the details of all datasets described in \cite{choe2020evaluating}.
In all three datasets, we select the best percentile for IVR from the \texttt{train-fullsup} set both in CUB and ImageNet.

\noindent
\textbf{Evaluation metric:}
In this paper, we use \texttt{MaxBox- AccV2} and \texttt{PxAP} as the evaluation metric as suggested in \cite{choe2020evaluating}.
For CUB and ImageNet, \texttt{MaxBoxAccV2}$(\delta)$ measures the performance in multiple intersection over union (IoU) $\delta \in \{0.3, 0.5, 0.7\}$ to address the trade-off between precision and recall.
\texttt{PxAP} measures the pixel-wise average precision when masks are available for evaluation.
Unlike preceding metrics, \texttt{MaxBoxAccV2}$(\delta)$ and \texttt{PxAP} choose the best operating threshold and report its score.
In this regard, we can make sure that any performance improvement made in this paper is not dependent on the choice of the threshold.
The evaluation has been conducted over VGG-GAP~\cite{simonyan2014very,zhou2016learning}, InceptionV3~\cite{szegedy2016rethinking}, and ResNet50~\cite{he2016deep}.

\begin{table*}[ht]
\caption{Evaluating WSOL using MaxBoxAccV2 with different normalization methods.}\smallskip
\centering
\setlength{\tabcolsep}{1.8pt}
\smallskip\begin{tabular}{|c|c|cccc|cccc|cccc|cccc|}
\Xhline{3\arrayrulewidth}
\multirow{2}{*}{Method} &
\multirow{2}{*}{Norm} &\multicolumn{4}{c|}{ImageNet (MaxBoxAccV2)} & \multicolumn{4}{c|}{CUB (MaxBoxAccV2)} & \multicolumn{4}{c|}{OpenImages (PxAP)}\\
\cline{3-14}
&  & VGG & Inception & ResNet & Mean & VGG & Inception & ResNet & Mean & VGG & Inception & ResNet & Mean \\
\hline\hline
\multirow{4}{*}{CAM} & Minmax & 60.02 & 63.40 & 63.65 & 62.36 & 63.71 & 56.68 & 62.98 & 61.13 & 58.30 & 63.23 & 58.49 & 60.01 \\
& Max & 60.85 & 64.77 & 64.88 & 63.50 & 64.54 & 59.10 & 63.47 & 62.37 & 59.65 & 64.97 & 59.49 & \textbf{61.37}\\
& PaS & 61.77 & 64.20 & 64.72 & 63.57 & 63.16 & 55.07 & 62.67 & 60.23 & 55.94 & 60.33 & 55.37 & 57.21\\
& IVR & 61.47 & 65.49 & 65.57 & \textbf{64.18} & 65.27 & 60.76 & 66.83 & \textbf{64.29} & 59.25 & 63.66 & 58.97 & 60.62\\
\cline{1-14}
\multirow{4}{*}{HaS} & Minmax & 60.59 & 63.72 & 63.40 & 62.57 & 63.71 & 53.38 & 64.63 & 60.58 & 58.14 & 58.11 & 55.93 & 57.39\\
& Max & 61.16 & 65.07 & 64.64 & 63.62 & 69.83 & 58.46 & 67.84 & 65.37 & 59.21 & 62.06 & 56.28 & \textbf{59.18}\\
& PaS & 62.05 & 64.53 & 64.57 & 63.72 & 61.16 & 51.27 & 61.85 & 58.09 & 55.97 & 57.79 & 52.85 & 55.54\\
& IVR & 61.62 & 65.71 & 64.90 & \textbf{64.08} & 71.77 & 60.56 & 71.23 & \textbf{67.85} & 58.78 & 61.31 & 55.68 & 58.59\\
\cline{1-14}
\multirow{4}{*}{ACoL} & Minmax & 57.43 & 63.69 & 62.29 & 61.14 & 57.38 & 56.18 & 66.43 & 59.99 & 54.34 & 57.02 & 57.25 & 56.26\\
& Max & 57.17 & 63.55 & 62.14 & 60.95 & 56.82 & 55.78 & 66.45 & 59.68 & 53.99 & 56.83 & 56.22 & 55.74\\
& PaS & 58.10 & 63.75 & 62.70 & 61.51 & 57.29 & 56.02 & 66.38 & 59.90 & 51.29 & 52.71 & 52.43 & 52.14\\
& IVR & 57.96 & 64.76 & 61.95 & \textbf{61.55} & 60.22 & 58.78 & 66.33 & \textbf{61.78} & 54.13 & 57.33 & 59.54 &\textbf{57.00} \\
\cline{1-14}
\multirow{4}{*}{SPG} & Minmax & 59.92 & 63.27 & 63.27 & 62.15 & 56.28 & 55.91 & 60.37 & 57.52 & 58.31 & 62.31 & 56.71 & 59.11\\
& Max & 60.44 & 64.63 & 64.04 & 63.04 & 60.21 & 56.66 & 60.85 & 59.24 & 59.20 & 64.50 & 57.42 & \textbf{60.37}\\
& PaS & 61.20 & 63.96 & 64.32 & 63.16 & 55.01 & 54.74 & 58.77 & 56.17 & 55.67 & 60.08 & 54.10 & 56.62\\
& IVR & 60.86 & 65.49 & 64.59 & \textbf{63.65} & 60.22 & 58.41 & 66.56 & \textbf{61.73} & 58.79 & 64.08 & 56.73 & 59.87\\
\cline{1-14}
\multirow{4}{*}{ADL} & Minmax & 59.86 & 61.43 & 63.65 & 61.65 & 66.27 & 58.81 & 58.33 & 61.14 & 58.65 & 56.85 & 55.15 & 56.88\\
& Max & 63.20 & 62.88 & 64.59 & 63.56 & 67.10 & 59.92 & 63.17 & 63.40 & 59.87 & 57.61 & 55.76 & \textbf{57.75} \\
& PaS & 60.86 & 62.30 & 64.49 & 62.55 & 64.97 & 57.32 & 56.67 & 59.65 & 55.99 & 54.42 & 53.06 & 54.49\\
& IVR & 63.71 & 64.01 & 65.06 & \textbf{64.26} & 67.25 & 60.28 & 64.30 & \textbf{63.94} & 59.25 & 56.88 & 55.03 & 57.05\\
\cline{1-14}
\multirow{4}{*}{CutMix} & Minmax & 59.46 & 63.86 & 63.27 & 62.20 & 62.31 & 57.43 & 62.76 & 60.83 & 58.07 & 62.56 & 57.73 & 59.45\\
& Max & 60.14 & 65.43 & 64.60 & 63.39 & 69.03 & 59.79 & 63.61 & 64.14 & 59.51 & 64.63 & 59.82 & \textbf{61.32}\\
& PaS & 61.07 & 64.40 & 64.42 & 63.30 & 60.94 & 55.90 & 62.58 & 59.90 & 55.72 & 60.12 & 55.47 & 57.10\\
& IVR & 60.57 & 66.04 & 64.93 & \textbf{63.84} & 71.54 & 61.74 & 67.14 & \textbf{66.80} & 58.89 & 63.24 & 58.95 & 60.36\\
\Xhline{3\arrayrulewidth}
\end{tabular}
\label{tab:wsolperf}
\end{table*}

\subsection{Analysis on the class activation map}
\noindent
$\textbf{Qualitative analysis:}$
Fig.~\ref{fig:camfeature} shows the class activation map and the region with values above the operating threshold $\tau^*_{30}$, $\tau^*_{50}$ and $\tau^*_{70}$ (red, green and blue respectively).
Note that $\tau^*_{70}$ is usually smaller than $\tau^*_{30}$ and $\tau^*_{50}$, capturing wider area in the images.
In all methods, min-max normalization and PaS suffer from mishandled class activation maps.
A region with exceptionally small values in the score map, which looks like a hole in black, raises the overall scores of all other pixels and makes the background more brighter than that of max normalization and IVR.
Especially for PaS, values in the background is even more higher than in min-max normalization.
Quite a lot of images possess sinkhole values in the class activation map and the optimal threshold becomes too low for them.
Consequently, the bounding box simply holds the whole image leading to a performance collapse.
Meanwhile, using max normalization successfully leave the background behind.
Negative values are simply ignored in the class activation map and a consistent threshold can be applied to any image.
Our IVR shows a similar aspect, localizing the object tightly in all three IoU thresholds.

Even though the activated area in the class activation map $\mathbf{F}$ is clearly visible in all normalization methods, we can understand how harmful the low values are in $\mathbf{F'}$.
If not processed properly, the performance of any method will no longer be evaluated right no matter how well the feature is trained.

\begin{figure*}[ht]
     \centering
     \begin{subfigure}[b]{0.33\textwidth}
         \includegraphics[width=\textwidth]{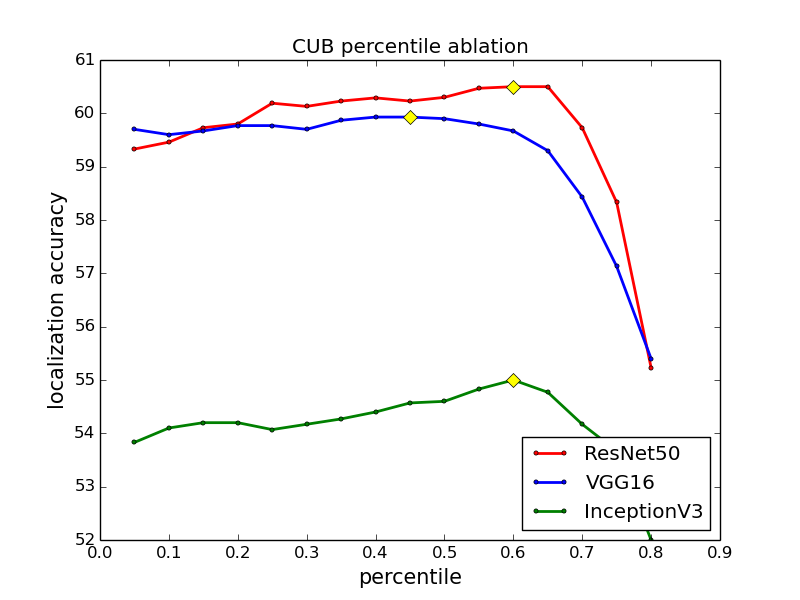}
          \caption{CUB}
         \label{fig:cam_percentile}
     \end{subfigure}
     \begin{subfigure}[b]{0.33\textwidth}
         \includegraphics[width=\textwidth]{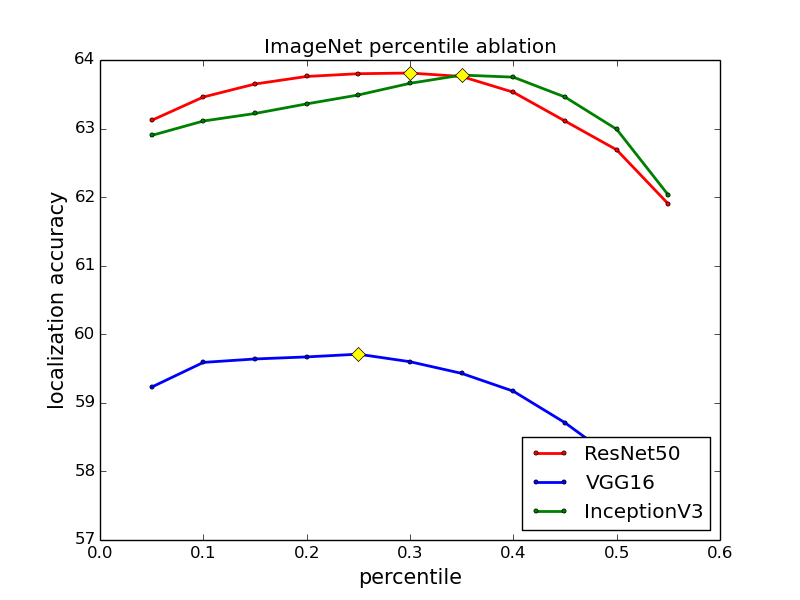}
          \caption{ImageNet}
         \label{fig:three sin x}
     \end{subfigure}
     \begin{subfigure}[b]{0.33\textwidth}
         \centering
         \includegraphics[width=\textwidth]{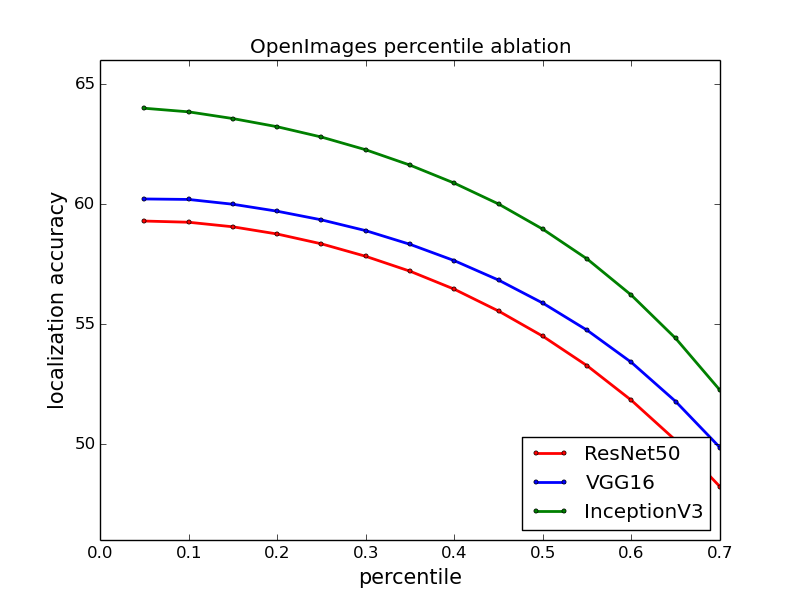}
          \caption{OpenImages}
         \label{fig:five over x}
     \end{subfigure}
        \caption{Localization accuracy measured with different percentile values when using IVR. The evaluation has been done only with CAM in the validation set of every datasets. To keep simplicity, the best percentile value in each architecture and dataset has been used in all other WSOL methods.}
        \label{fig:percentile_ablation}
\end{figure*}

\noindent
$\textbf{Quantitative analysis: }$
The significance of normalization can be verified numerically as well.
Fig.~\ref{fig:minmaxdist} shows the distribution of minimum and maximum values within the class activation maps of all test images.
Note that values expressed in Fig.~\ref{fig:minmaxdist} are those before normalization and the localization are done after each normalization method.
HaS with VGG16 has been used for the comparison in CUB and ADL with VGG16 has been used for ImageNet.
The minimum values are plotted on the x-axis while maximum values are plotted on the y-axis.
In other words, a single point can express the values range in class activation map $\mathbf{F}$ of an image.
Applying the optimal threshold $\tau^*$ of IoU 70, black and red dots represent positive and negative samples of localization.
In both datasets, min-max normalization shows the most inefficient shape of distribution.
Especially, the variation of maximum values does not seem to affect the localization performance in a great deal.
Meanwhile, numerous number of samples with highly negative minimum values fail to be localized.
Even though the scale of minimum values are much smaller than that of maximum values, we can see that the variation in minimum values makes the model harder to localize objects.
This can be proven in that almost all samples with a minimum value under -0.07 fail in localization in CUB.
In the graph of max normalization, much more samples have successfully been localized.
The distribution of positive and negative samples seems to be evenly distributed.
In this point, we can assume that highly negative values in the class activation map are unnecessary for localization.
The area covered by PaS is a little bit wider than that of min-max normalization but still it does not completely restrain the influence of extremely small minimum values.
As explained in the previous sections, IVR can be assumed as a variant of max normalization and the distribution resembles that of max normalization.

\subsection{Evaluation of WSOL methods}
All hyperparameters pioneered by \cite{choe2020evaluating} are available online\footnote{https://github.com/clovaai/wsolevaluation}.
Using these optimized values obtained from a hyperparameter search, we can avoid issues coming from selecting different hyperparameters.
WSOL methods, architectures, and datasets are the same as in \cite{choe2020evaluating} as well.

Tab.~\ref{tab:wsolperf} shows the performances of all concerned experiments.
In ImageNet, min-max normalization shows the worst performance in all WSOL methods except ACoL.
With min-max normalization, CAM still works better than all the other WSOL methods.
When using max normalization, all methods except ACoL improve from min-max normalization slightly.
In this case, ADL performs the best for ImageNet.
A little bit higher overall performance than max normalization can be seen in PaS except ADL.
Note that ADL with PaS is still better than min-max normalization.
IVR has enhanced all WSOL methods without any exception.
In this case, ADL has shown the best performance when all architectures are averaged.
When measured individually, CutMix with IVR has recorded 66.04\% in Inception which is the best ImageNet performance in total.

In CUB, PaS shows lower performance than min-max normalization in all WSOL methods.
Note that the reported scores in this paper are higher than those in \cite{bae2020rethinking}.
For example, \cite{bae2020rethinking} reports 65.90 of IoU 50 localization accuracy in CAM with PaS.
In our experiment, IoU 50 accuracy of CAM with PaS has recorded 71.61 since much more better hyperparameters from \cite{choe2020evaluating} are available.
Meanwhile, IVR has shown a drastic improvement in all experiments.
Especially, HaS and CutMix have exceeded 70\% in VGG, which has never been reached by any other normalization methods.
Unlike in ImageNet, the performance gap among WSOL methods becomes considerably noticeable.

In OpenImages, the performance of PaS rather decreases and even the performance of IVR is slightly better than min-max normalization. 
Max normalization shows an overall improvement and CAM is still the best method.
IVR is the second best normalization method while PaS is the worst one.
Even with \texttt{PxAP}, low values rather than high values affect every pixels in the class activation map leading to a performance degradation.
Consequently, CAM still works better than or at least almost equal to the best method in both ImageNet and OpenImages. 
However, it has yielded its top position to HaS in CUB.
The conclusion of \cite{choe2020evaluating} that many CAM-based WSOL methods have barely improved the original CAM is correct only in ImageNet and OpenImages.

We have also conducted experiments using PaS and IVR simultaneously.
In all cases, the resulting score is much better than using PaS only but slightly worse than using IVR alone.
Therefore, we omit the result in the table to avoid redundancy.

\subsection{Percentile selection in IVR}
In PaS, the choice of the maximum value percentile has been made empirically.
To alleviate the ambiguity of choosing a percentile for IVR, we have used the validation set and chosen the best values.
Fig.~\ref{fig:percentile_ablation} shows how the percentile value affects the performance in CUB, ImageNet, and OpenImages.

In CUB, class activation maps extracted from images have shown a large variation in minimum values as described in Fig.~\ref{fig:minmaxdist}.
Consistently, removing much of the portion from an activation map has shown better performance.
Using 45-th, 60-th, and 60-th percentile have peaked in VGG, ResNet, and Inception respectively.
In all cases, these values are all positive and this implies that even positive values in the class activation map may not contribute to the localization performance.
In ImageNet, 25-th, 30-th, and 35-th percentile work best in VGG, ResNet and Inception respectively.
In other words, ImageNet uses a wider range in minimum value distribution than CUB.
As in Fig.~\ref{fig:minmaxdist}, this is because the variation of minimum values in ImageNet is relatively smaller than that in CUB.
In OpenImages, using IVR has constantly degraded the performance from 0-th percentile in all architectures.
Experiments in Tab.~\ref{tab:wsolperf} uses 5-th percentile in all the cases.

\begin{table}[t]
\caption{Localization accuracy when using Negative Weight Clamping additionally \cite{bae2020rethinking}.}\smallskip
\centering
\setlength{\tabcolsep}{5pt}
\smallskip\begin{tabular}{c|c|c|c|c|c}
\Xhline{3\arrayrulewidth}
\multirow{2}{*}{Method} & \multirow{2}{*}{Norm} & \multicolumn{4}{c}{Localization accuracy}\\
\cline{3-6}
& & IoU 30 & IoU 50 & IoU 70 & Mean\\
\hline\hline
\multirow{4}{*}{HaS} & Min-max & 97.62 & 84.43 & 43.89 & 75.31\\
& Max & 97.45 & 83.88 & 43.36 & 74.89\\
& PaS & 94.74 & 75.47 & 39.75 & 69.99\\
& IVR & 98.41 & 86.56 & 46.82 & \textbf{77.26}\\
\hline
\multirow{4}{*}{CutMix} & Min-max & 98.50 & 88.13 & 48.03 & 78.22\\
& Max & 98.39 & 87.94 & 47.53 & 77.95\\
& PaS & 97.39 & 82.91 & 42.87 & 74.39\\
& IVR & 98.74 & 89.25 & 49.59 & \textbf{79.19}\\
\Xhline{3\arrayrulewidth}
\end{tabular}
\label{tab:nwc}
\end{table}


\noindent
\textbf{Evaluation with NWC~\cite{bae2020rethinking}}
In the previous sections, we have mainly discussed about the normalization of class activation maps and excluded the usage of Thresholded Average Pooling (TAP) and Negative Weight Clamping (NWC) proposed in \cite{bae2020rethinking} when using PaS.
TAP alleviates the problem that the global averaging pooling layer does not reflect the difference in different channels.
By applying a threshold during average pooling, the model can focus on important activations in each channel while training.
As this is a technique used in training time, we do not discuss about it in this paper.
Meanwhile, NWC is a method presuming that negative weights from the classifier does not contribute to the localization.
Negative weights are clamped to zero and low values in the class activation map may be excluded.
The authors of \cite{bae2020rethinking} claim that PaS improves CAM-based WSOL methods by itself but verification of other methods with NWC is necessary.
Tab.~\ref{tab:nwc} shows the comparison of all normalization methods combined with NWC in CUB.
Both in HaS and CutMix, PaS still shows the worst performance.
Even considering the best localization accuracy 78.58\% of HaS at IoU 50 reported in \cite{bae2020rethinking}, scores of other normalization methods exceed at least 83\%.
IVR in CutMix has scored 79.19\% in average and especially 89.25\% at IoU 50.
To the best of our knowledge, this is the best localization score reported in CUB.

\section{Discussion}
\begin{figure}[t]
\begin{center}
  \includegraphics[width=\linewidth]{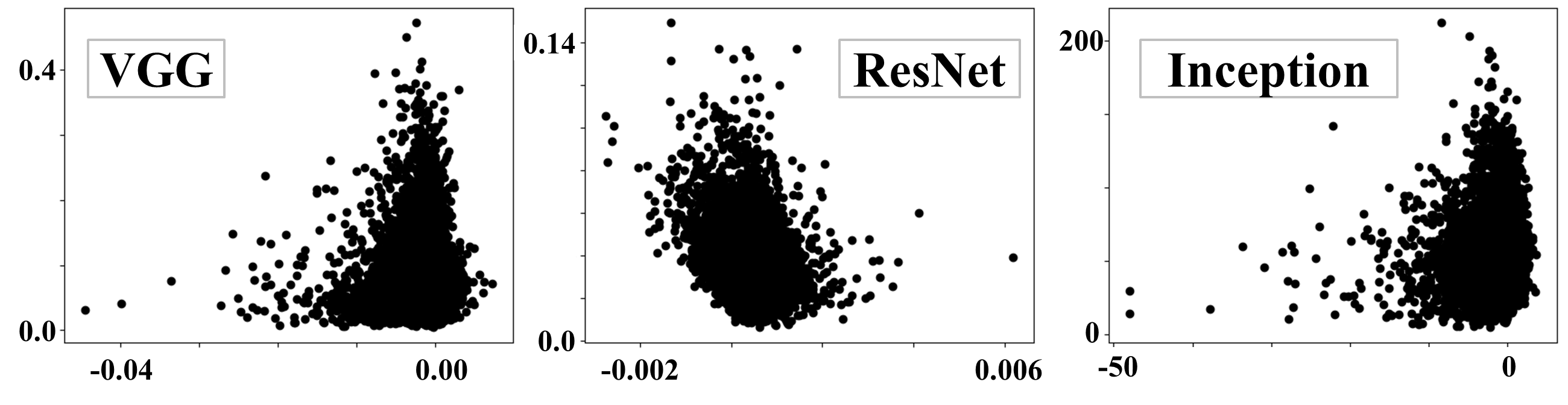}
\end{center}
  \caption{Distribution of minimum and maximum values from class activation map $F$ of all test images in OpenImages. Compared to CUB and ImageNet, the distribution of maximum values is more influential.}
\label{fig:openimage_dist}
\end{figure}

In experiments above, max normalization shows the best result in OpenImages.
This can be attributed to the property of the dataset.
Fig.~\ref{fig:openimage_dist} shows the minimum and maximum values of CAM with three different architectures' class activation maps in OpenImages.
As in the figure, the variation of minimum values is much smaller than those of CUB and ImageNet.
Numerically, the ratio of the standard deviation of all maximum values to the one of all minimum values in VGG has recorded approximately 11, 12.04, and 18.36 in CUB, ImageNet, and OpenImages respectively.
In case of ResNet, this ratio has recorded up to 31.79 and the performance improvement of max normalization and IVR in ResNet is insignificant as in VGG.
Also, Inception has recorded 8.19 and the performance improvement is better than other architectures.
Depending on the dataset and architecture, we can see that this rate and the optimal percentile are inversely proportional.
When the variation of maximum values is much more higher, using max normalization instead of any other normalization is likely to be a better choice.
Therefore depending on the property of the dataset and architecture, we recommend to use an appropriate normalization method.

Also, all WSOL methods emerged after CAM have indeed improved CAM in CUB, contrary to \cite{choe2020evaluating}, while they have hardly contributed in ImageNet and OpenImages as concluded in \cite{choe2020evaluating}. 
Because CUB is for a fine-grained classification problem, all the images share common features of a bird. Therefore, a classifier for CUB dataset has to discriminate fine details and the common features have little discriminative power. Recent WSOL methods have focused on compensating these relatively less discriminative features.
According to our investigation, WSOL methods studied so far have successfully dealt with this problem.
On the other hand, they have not focused on more general classification problems such as ImageNet and OpenImages and there remains much room for investigation for this problem. 

\section{Conclusion}
For several years, an issue about whether many proposed weakly supervised object localization methods have actually derived improvement has been raised.
Meanwhile, there has been few works dealing with the normalization effect of class activation maps.
In this paper, we proposed a new and effective normalization method along with a solid evaluation with many other possible normalization methods.
The new proposed normalization method achieves the new state-of-the-art performance in CUB-200-2011 and ImageNet dataset.
Also, we point out that the normalization method should be selected according to the traits of the dataset.

For future works in WSOL, we suggest that even though many WSOL methods successfully improved the performance in a dataset like CUB, a new perspective which will also work in real world datasets such as ImageNet and OpenImages is still in need.

\section*{Acknowledgements}
This work was supported by NRF grant (2021R1A2C3 006659) and IITP grant (2021-0-00537), both funded by the Government of the Republic of Korea.


{\small
\bibliographystyle{ieee_fullname}
\bibliography{egbib}

\begin{thebibliography}{10}\itemsep=-1pt

\bibitem{bae2020rethinking}
Wonho Bae, Junhyug Noh, and Gunhee Kim.
\newblock Rethinking class activation mapping for weakly supervised object
  localization.
\newblock In {\em European Conference on Computer Vision}, pages 618--634.
  Springer, 2020.

\bibitem{bergstra2012random}
James Bergstra and Yoshua Bengio.
\newblock Random search for hyper-parameter optimization.
\newblock {\em Journal of machine learning research}, 13(2), 2012.

\bibitem{bochkovskiy2020yolov4}
Alexey Bochkovskiy, Chien-Yao Wang, and Hong-Yuan~Mark Liao.
\newblock Yolov4: Optimal speed and accuracy of object detection.
\newblock {\em arXiv preprint arXiv:2004.10934}, 2020.

\bibitem{chattopadhay2018grad}
Aditya Chattopadhay, Anirban Sarkar, Prantik Howlader, and Vineeth~N
  Balasubramanian.
\newblock Grad-cam++: Generalized gradient-based visual explanations for deep
  convolutional networks.
\newblock In {\em 2018 IEEE Winter Conference on Applications of Computer
  Vision (WACV)}, pages 839--847. IEEE, 2018.

\bibitem{chen2018encoder}
Liang-Chieh Chen, Yukun Zhu, George Papandreou, Florian Schroff, and Hartwig
  Adam.
\newblock Encoder-decoder with atrous separable convolution for semantic image
  segmentation.
\newblock In {\em Proceedings of the European conference on computer vision
  (ECCV)}, pages 801--818, 2018.

\bibitem{choe2020evaluating}
Junsuk Choe, Seong~Joon Oh, Seungho Lee, Sanghyuk Chun, Zeynep Akata, and
  Hyunjung Shim.
\newblock Evaluating weakly supervised object localization methods right.
\newblock In {\em Proceedings of the IEEE/CVF Conference on Computer Vision and
  Pattern Recognition}, pages 3133--3142, 2020.

\bibitem{choe2019attention}
Junsuk Choe and Hyunjung Shim.
\newblock Attention-based dropout layer for weakly supervised object
  localization.
\newblock In {\em Proceedings of the IEEE/CVF Conference on Computer Vision and
  Pattern Recognition}, pages 2219--2228, 2019.

\bibitem{fu2019adaptive}
Jun Fu, Jing Liu, Yuhang Wang, Yong Li, Yongjun Bao, Jinhui Tang, and Hanqing
  Lu.
\newblock Adaptive context network for scene parsing.
\newblock In {\em Proceedings of the IEEE/CVF International Conference on
  Computer Vision}, pages 6748--6757, 2019.

\bibitem{he2016deep}
Kaiming He, Xiangyu Zhang, Shaoqing Ren, and Jian Sun.
\newblock Deep residual learning for image recognition.
\newblock In {\em Proceedings of the IEEE conference on computer vision and
  pattern recognition}, pages 770--778, 2016.

\bibitem{OpenImages}
Alina Kuznetsova, Hassan Rom, Neil Alldrin, Jasper Uijlings, Ivan Krasin, Jordi
  Pont-Tuset, Shahab Kamali, Stefan Popov, Matteo Malloci, Alexander
  Kolesnikov, Tom Duerig, and Vittorio Ferrari.
\newblock The open images dataset v4: Unified image classification, object
  detection, and visual relationship detection at scale.
\newblock {\em IJCV}, 2020.

\bibitem{lin2019zigzagnet}
Di Lin, Dingguo Shen, Siting Shen, Yuanfeng Ji, Dani Lischinski, Daniel
  Cohen-Or, and Hui Huang.
\newblock Zigzagnet: Fusing top-down and bottom-up context for object
  segmentation.
\newblock In {\em Proceedings of the IEEE/CVF Conference on Computer Vision and
  Pattern Recognition}, pages 7490--7499, 2019.

\bibitem{lin2017focal}
Tsung-Yi Lin, Priya Goyal, Ross Girshick, Kaiming He, and Piotr Doll{\'a}r.
\newblock Focal loss for dense object detection.
\newblock In {\em Proceedings of the IEEE international conference on computer
  vision}, pages 2980--2988, 2017.

\bibitem{liu2016ssd}
Wei Liu, Dragomir Anguelov, Dumitru Erhan, Christian Szegedy, Scott Reed,
  Cheng-Yang Fu, and Alexander~C Berg.
\newblock Ssd: Single shot multibox detector.
\newblock In {\em European conference on computer vision}, pages 21--37.
  Springer, 2016.

\bibitem{recht2019imagenet}
Benjamin Recht, Rebecca Roelofs, Ludwig Schmidt, and Vaishaal Shankar.
\newblock Do imagenet classifiers generalize to imagenet?
\newblock In {\em International Conference on Machine Learning}, pages
  5389--5400. PMLR, 2019.

\bibitem{redmon2016you}
Joseph Redmon, Santosh Divvala, Ross Girshick, and Ali Farhadi.
\newblock You only look once: Unified, real-time object detection.
\newblock In {\em Proceedings of the IEEE conference on computer vision and
  pattern recognition}, pages 779--788, 2016.

\bibitem{ren2015faster}
Shaoqing Ren, Kaiming He, Ross Girshick, and Jian Sun.
\newblock Faster r-cnn: Towards real-time object detection with region proposal
  networks.
\newblock {\em arXiv preprint arXiv:1506.01497}, 2015.

\bibitem{russakovsky2015imagenet}
Olga Russakovsky, Jia Deng, Hao Su, Jonathan Krause, Sanjeev Satheesh, Sean Ma,
  Zhiheng Huang, Andrej Karpathy, Aditya Khosla, Michael Bernstein, et~al.
\newblock Imagenet large scale visual recognition challenge.
\newblock {\em International journal of computer vision}, 115(3):211--252,
  2015.

\bibitem{samek2016evaluating}
Wojciech Samek, Alexander Binder, Gr{\'e}goire Montavon, Sebastian Lapuschkin,
  and Klaus-Robert M{\"u}ller.
\newblock Evaluating the visualization of what a deep neural network has
  learned.
\newblock {\em IEEE transactions on neural networks and learning systems},
  28(11):2660--2673, 2016.

\bibitem{selvaraju2017grad}
Ramprasaath~R Selvaraju, Michael Cogswell, Abhishek Das, Ramakrishna Vedantam,
  Devi Parikh, and Dhruv Batra.
\newblock Grad-cam: Visual explanations from deep networks via gradient-based
  localization.
\newblock In {\em Proceedings of the IEEE international conference on computer
  vision}, pages 618--626, 2017.

\bibitem{simonyan2013deep}
Karen Simonyan, Andrea Vedaldi, and Andrew Zisserman.
\newblock Deep inside convolutional networks: Visualising image classification
  models and saliency maps.
\newblock {\em arXiv preprint arXiv:1312.6034}, 2013.

\bibitem{simonyan2014very}
Karen Simonyan and Andrew Zisserman.
\newblock Very deep convolutional networks for large-scale image recognition.
\newblock {\em arXiv preprint arXiv:1409.1556}, 2014.

\bibitem{singh2017hide}
Krishna~Kumar Singh and Yong~Jae Lee.
\newblock Hide-and-seek: Forcing a network to be meticulous for
  weakly-supervised object and action localization.
\newblock In {\em 2017 IEEE international conference on computer vision
  (ICCV)}, pages 3544--3553. IEEE, 2017.

\bibitem{szegedy2016rethinking}
Christian Szegedy, Vincent Vanhoucke, Sergey Ioffe, Jon Shlens, and Zbigniew
  Wojna.
\newblock Rethinking the inception architecture for computer vision.
\newblock In {\em Proceedings of the IEEE conference on computer vision and
  pattern recognition}, pages 2818--2826, 2016.

\bibitem{tan2020efficientdet}
Mingxing Tan, Ruoming Pang, and Quoc~V Le.
\newblock Efficientdet: Scalable and efficient object detection.
\newblock In {\em Proceedings of the IEEE/CVF conference on computer vision and
  pattern recognition}, pages 10781--10790, 2020.

\bibitem{welinder2010caltech}
Peter Welinder, Steve Branson, Takeshi Mita, Catherine Wah, Florian Schroff,
  Serge Belongie, and Pietro Perona.
\newblock Caltech-ucsd birds 200.
\newblock 2010.

\bibitem{yun2019cutmix}
Sangdoo Yun, Dongyoon Han, Seong~Joon Oh, Sanghyuk Chun, Junsuk Choe, and
  Youngjoon Yoo.
\newblock Cutmix: Regularization strategy to train strong classifiers with
  localizable features.
\newblock In {\em Proceedings of the IEEE/CVF International Conference on
  Computer Vision}, pages 6023--6032, 2019.

\bibitem{zhang2020bridging}
Shifeng Zhang, Cheng Chi, Yongqiang Yao, Zhen Lei, and Stan~Z Li.
\newblock Bridging the gap between anchor-based and anchor-free detection via
  adaptive training sample selection.
\newblock In {\em Proceedings of the IEEE/CVF Conference on Computer Vision and
  Pattern Recognition}, pages 9759--9768, 2020.

\bibitem{zhang2018adversarial}
Xiaolin Zhang, Yunchao Wei, Jiashi Feng, Yi Yang, and Thomas~S Huang.
\newblock Adversarial complementary learning for weakly supervised object
  localization.
\newblock In {\em Proceedings of the IEEE Conference on Computer Vision and
  Pattern Recognition}, pages 1325--1334, 2018.

\bibitem{zhang2018self}
Xiaolin Zhang, Yunchao Wei, Guoliang Kang, Yi Yang, and Thomas Huang.
\newblock Self-produced guidance for weakly-supervised object localization.
\newblock In {\em Proceedings of the European conference on computer vision
  (ECCV)}, pages 597--613, 2018.

\bibitem{zhao2017pyramid}
Hengshuang Zhao, Jianping Shi, Xiaojuan Qi, Xiaogang Wang, and Jiaya Jia.
\newblock Pyramid scene parsing network.
\newblock In {\em Proceedings of the IEEE conference on computer vision and
  pattern recognition}, pages 2881--2890, 2017.

\bibitem{zhou2016learning}
Bolei Zhou, Aditya Khosla, Agata Lapedriza, Aude Oliva, and Antonio Torralba.
\newblock Learning deep features for discriminative localization.
\newblock In {\em Proceedings of the IEEE conference on computer vision and
  pattern recognition}, pages 2921--2929, 2016.

\bibitem{zhu2019improving}
Yi Zhu, Karan Sapra, Fitsum~A Reda, Kevin~J Shih, Shawn Newsam, Andrew Tao, and
  Bryan Catanzaro.
\newblock Improving semantic segmentation via video propagation and label
  relaxation.
\newblock In {\em Proceedings of the IEEE/CVF Conference on Computer Vision and
  Pattern Recognition}, pages 8856--8865, 2019.

\end{thebibliography}
}

\clearpage

\appendix
\section{Details about datasets}

In the work of \cite{choe2020evaluating}, Caltech-UCSD Birds-200-2011 (CUB) consists of 5994 "train" images and 5794 "test" images for 200 classes and they are used as the \texttt{train-weaksup} and \texttt{test} set respectively.
The \texttt{train-fullsup} has been additionally collected from Flickr with extra annotated bounding boxes and used as the validation set.
Similarly, 1.2M ``train" images and 10K ``validation" images for 1000 classes in ImageNet are used as the \texttt{train-weaksup} and \texttt{test} set respectively.
Images with annotated bounding boxes from the ImageNetV2 \cite{recht2019imagenet} are used as the \texttt{train-fullsup} set.
In both datasets, we select the best percentile for IVR from the \texttt{train-fullsup} set both in CUB and ImageNet.
Additionally, we use OpenImages~\cite{OpenImages} which is reorganized in \cite{welinder2010caltech} with 29819, 2500, and 5000 images of \texttt{train-weaksup, train-fullsup} and \texttt{test}.
Even though the evaluation metric in OpenImages is different from that of CUB and ImageNet, we still use \texttt{train-fullsup} as a validation set to find the optimal percentile for IVR.

\section{Visualization of localized samples}
\begin{figure*}[t]
\begin{center}
   \includegraphics[width=0.84\linewidth]{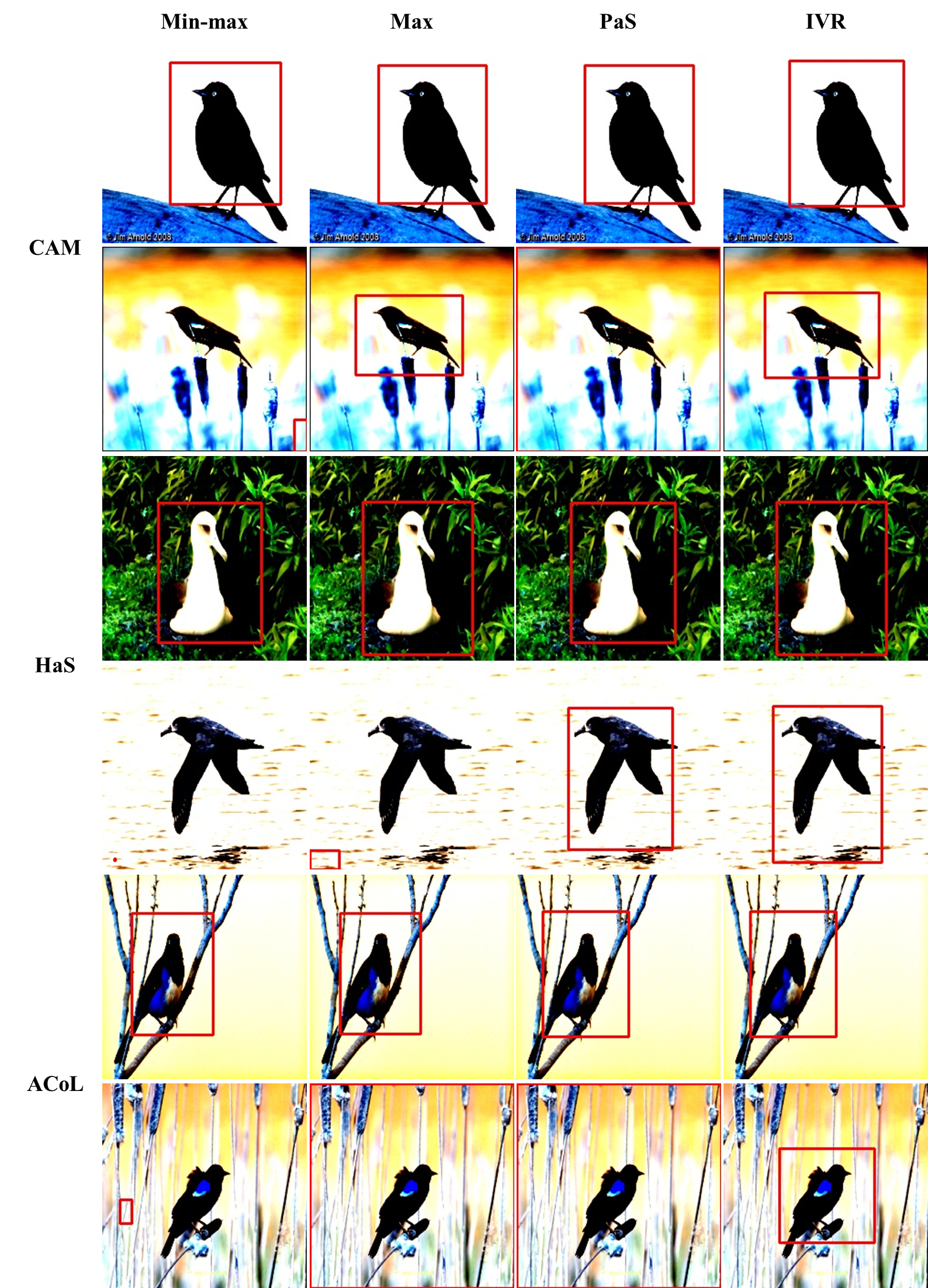}
\end{center}
   \caption{Localized examples of CAM, HaS and ACoL in CUB.}
\label{fig:cub_camhasacol}
\end{figure*}

\begin{figure*}[t]
\begin{center}
   \includegraphics[width=0.84\linewidth]{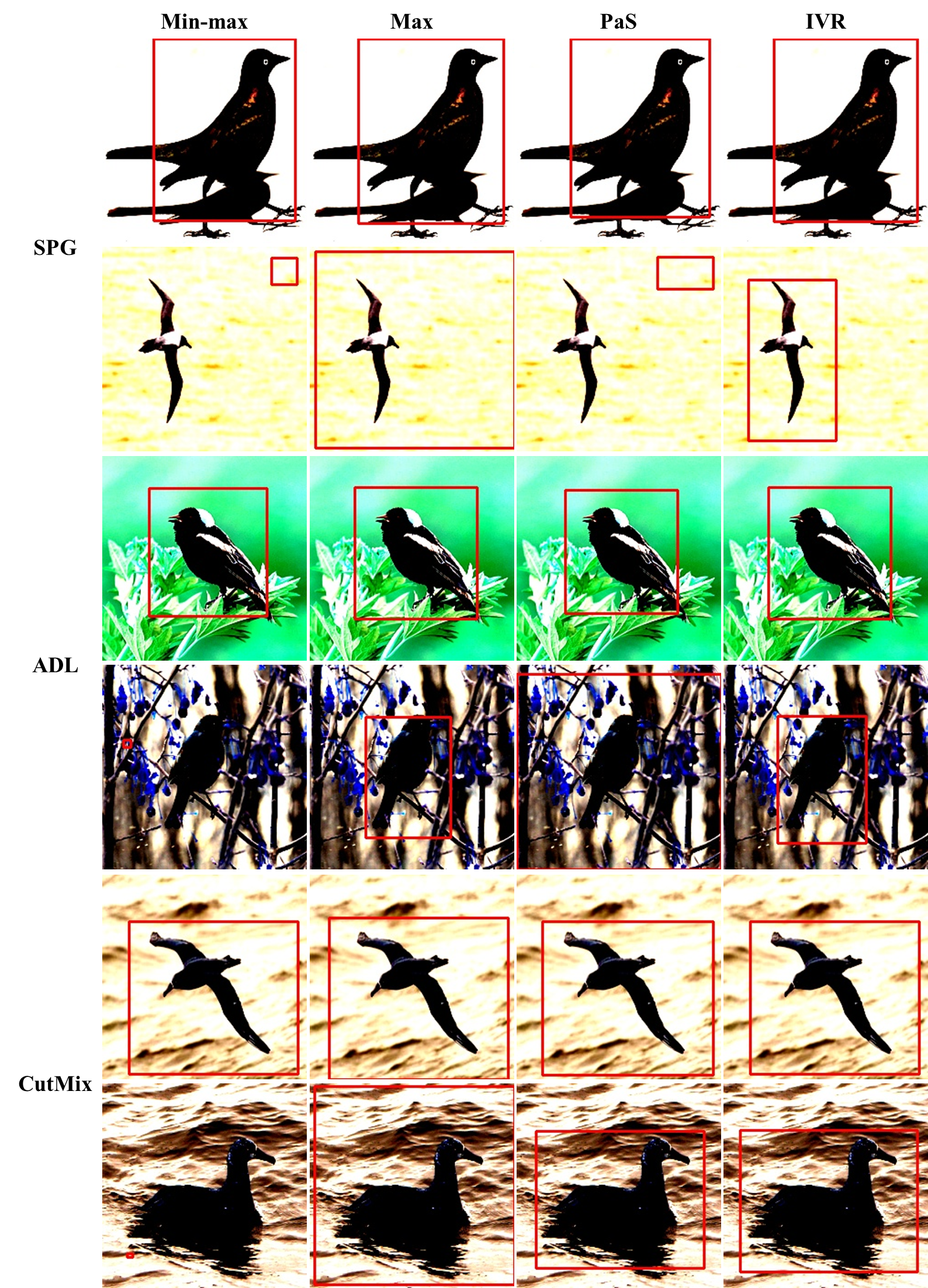}
\end{center}
   \caption{Localized examples of SPG, ADL and CutMix in CUB.}
\label{fig:cub_spgadlcutmix}
\end{figure*}

\begin{figure*}[t]
\begin{center}
   \includegraphics[width=0.84\linewidth]{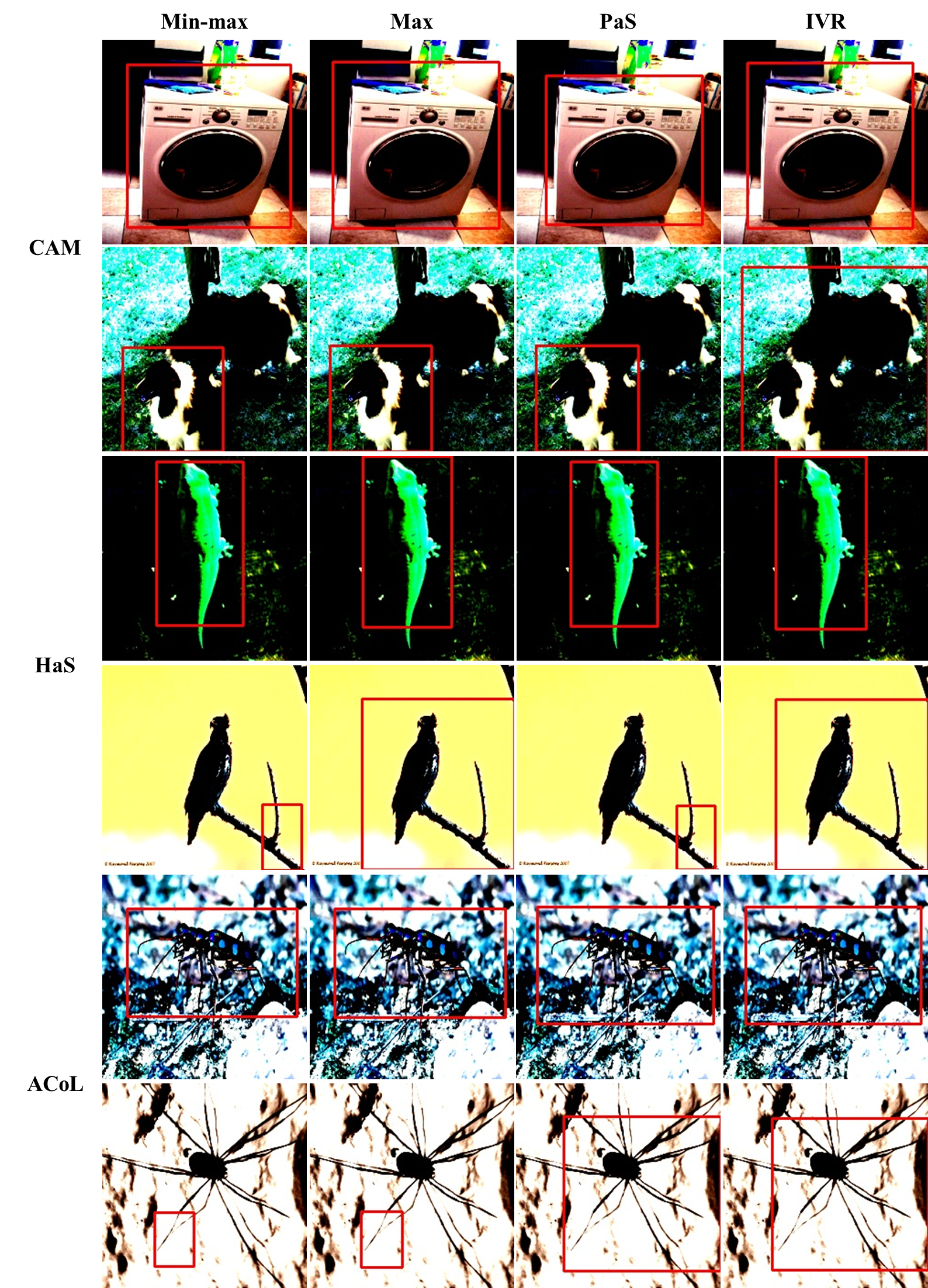}
\end{center}
   \caption{Localized examples of CAM, HaS and ACoL in ImageNet.}
\label{fig:imagenet_camhasacol}
\end{figure*}

\begin{figure*}[t]
\begin{center}
   \includegraphics[width=0.84\linewidth]{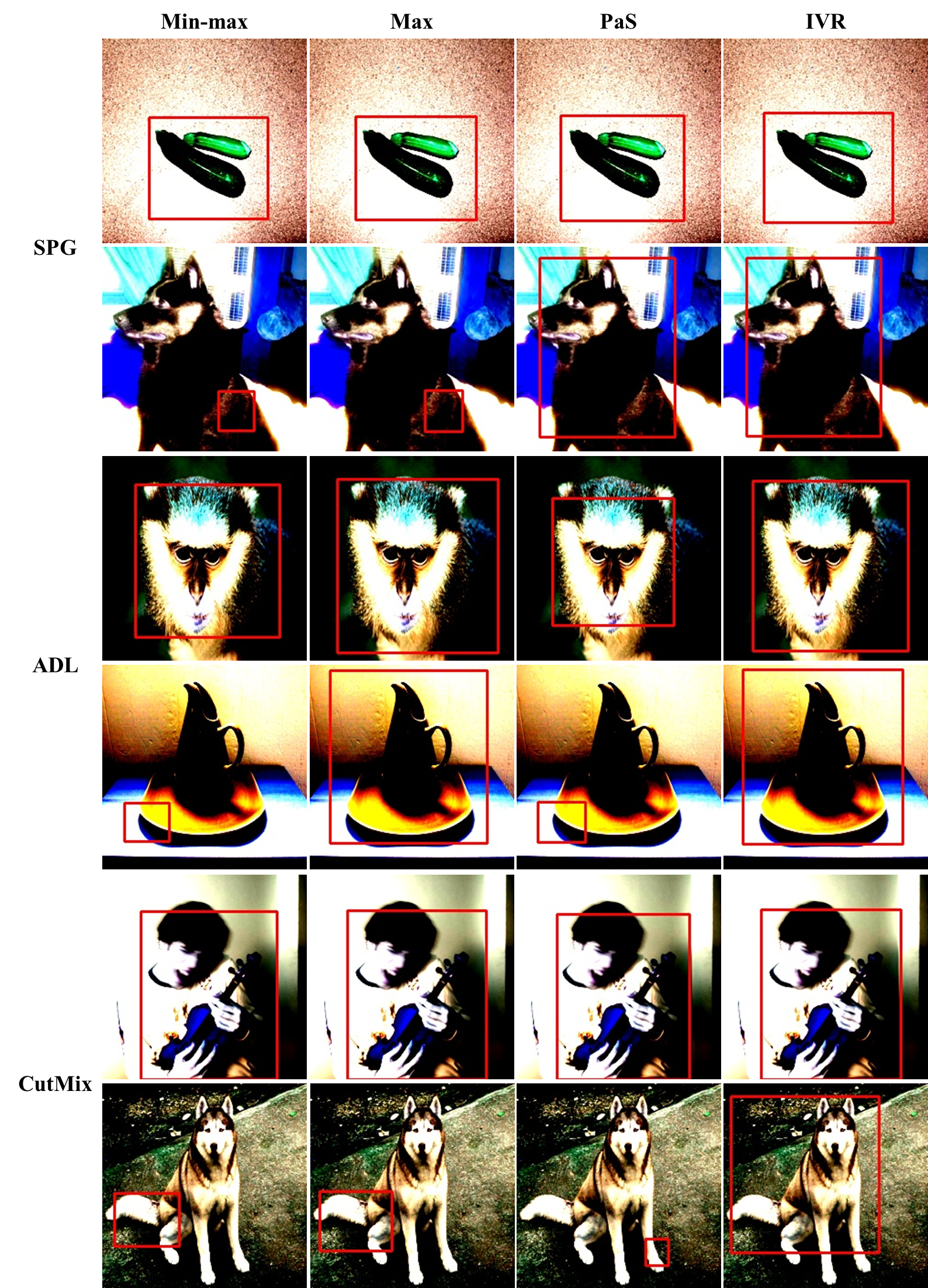}
\end{center}
   \caption{Localized examples of SPG, ADL and CutMix in ImageNet.}
\label{fig:imagenet_spgadlcutmix}
\end{figure*}

Fig.~\ref{fig:cub_camhasacol},\ref{fig:cub_spgadlcutmix} show the localization results in CUB and Fig.~\ref{fig:imagenet_camhasacol},\ref{fig:imagenet_spgadlcutmix} show the localization results in ImageNet.
The red boxes are bounding boxes upon the threshold of IoU 70.
In each method, correctly localized samples in all normalization methods are placed on the first row.
Using the same threshold, samples on the second row show some failure cases in each normalization method.
Even an optimal threshold searched throughout a dataset fails localization in many samples.

\section{Evaluation with Negative Weight Clamping (NWC)}
NWC is a very powerful method for WSOL as it prevents contradictory features from disturbing each other.
Tab.~\ref{tab:nwcperf} shows the results of all methods combined with NWC.
To find out the best performance score, IVR has used the best percentile verified from the validation set in all individual experiments.

In ImageNet, the performance improvement from min-max normalization is not significant, but PaS has shown the best result in CAM, HaS, ADL, and CutMix when combined with NWC.
Also, unlike the results without NWC, max normalization has been no more better than min-max normalization.
IVR is still better than min-max normalization and outperforms PaS only in ACoL and SPG.
Among all cases, the top localization accuracy in IoU 50 recorded 68.73\% in CutMix-Inception with PaS while the best score reported in \cite{bae2020rethinking} is 64.44\%.

In CUB, IVR outperforms all other methods in a great deal.
Min-max normalization is also comparable with max normalization.
Meanwhile, the performance in PaS significantly drops in all cases.
Using the individual best percentile acquired from the validation set, 89.14\% of localization accuracy in IoU 50 can be acquired in CutMix-VGG with IVR.

In OpenImages, PaS shows the greatest performance degradation in all cases.
Instead, contrary to the results in the main paper, min-max normalization shows the best result in OpenImages.
IVR remains in the second best normalization method with a slightly lower score compared to min-max normalization.

According to this result, removing negative values in the weight makes max normalization unnecessary.
However, IVR still shows comparable performance in ImageNet and OpenImages, and far more better performance in CUB.
Also in Tab.~\ref{tab:nwcperf}, we provide the performance drop from the top score among the four normalization methods in red.
The variation of these values among three datasets is reported in the last column and we can easily see that IVR with NWC shows the most stable and robust performance.
This implies that the minimum value in the class activation map must be selected adaptively.
Researchers should remember that each WSOL method requires a specific normalization method that fits best in different datasets.

\vfill

\begin{table*}[t]
\caption{Evaluating WSOL with different normalization methods with Negative Weight Clamping (NWC)~\cite{bae2020rethinking}. Numbers in red are the performance drop from the top score among the four normalization methods. The last column represents the variance of these scores in one normalization method. IVR shows the smallest variance.}\smallskip
\centering
\resizebox{\textwidth}{!}{
\setlength{\tabcolsep}{0.6pt}
\smallskip\begin{tabular}{c|c|cccc|cccc|cccc|cccc||c}
\Xhline{3\arrayrulewidth}
\multirow{2}{*}{Method} &
\multirow{2}{*}{Norm} &\multicolumn{4}{c|}{ImageNet (MaxBoxAccV2)} & \multicolumn{4}{c|}{CUB (MaxBoxAccV2)} & \multicolumn{4}{c|}{OpenImages (PxAP)} & \multirow{2}{*}{Var} \\
\cline{3-14}
&  & VGG & Incep. & ResNet & Mean & VGG & Incep. & ResNet & Mean & VGG & Incep. & ResNet & Mean &\\
\hline\hline
\multirow{4}{*}{CAM} & Minmax & 60.28 & 64.47 & 64.77 & 63.17(\textcolor{red}{-1.51}) & 68.90 & 62.39 & 69.26 & 66.85(\textcolor{red}{-0.60}) & 59.73 & 64.30 & 59.90 & \textbf{61.31}(\textcolor{red}{0.00}) & 0.58\\
& Max & 59.58 & 64.41 & 64.50 & 62.83(\textcolor{red}{-1.85}) & 68.82 & 62.24 & 68.58 & 66.54(\textcolor{red}{-0.91}) & 59.20 & 64.44 & 59.87 & 61.17(\textcolor{red}{-0.14}) & 0.74\\
& PaS & 62.45 & 65.21 & 66.38 & \textbf{64.68}(\textcolor{red}{0.00}) & 67.86 & 60.51 & 68.66 & 65.68(\textcolor{red}{-1.78}) & 55.96 & 59.69 & 55.42 & 57.02(\textcolor{red}{-4.29}) & 4.64\\
& IVR & 60.98 & 65.32 & 65.51 & 63.94(\textcolor{red}{-0.74}) & 68.95 & 62.83 & 70.57 & \textbf{67.45}(\textcolor{red}{0.00}) & 59.66 & 63.88 & 59.57 & 61.04(\textcolor{red}{-0.27}) & \textbf{0.20}\\
\cline{1-15}
\multirow{4}{*}{HaS} & Minmax & 61.16 & 64.40 & 64.72 & 63.43(\textcolor{red}{-1.39}) & 75.31 & 61.74 & 74.48 & 70.51(\textcolor{red}{-2.16}) & 59.16 & 62.73 & 56.67 & \textbf{59.52}(\textcolor{red}{0.00}) & 1.20\\
& Max & 60.75 & 64.35 & 64.47 & 63.19(\textcolor{red}{-1.63}) & 74.89 & 62.02 & 74.01 & 70.31(\textcolor{red}{-2.37}) & 58.58 & 62.67 & 56.78 & 59.34(\textcolor{red}{-0.18}) & 1.24\\
& PaS & 62.89 & 65.36 & 66.21 & \textbf{64.82}(\textcolor{red}{0.00}) & 69.99 & 60.17 & 71.20 & 67.12(\textcolor{red}{-5.56}) & 55.95 & 57.11 & 53.21 & 55.42(\textcolor{red}{-4.10}) & 8.30\\
& IVR & 61.56 & 65.22 & 65.16 & 63.98(\textcolor{red}{-0.84}) & 77.59 & 64.15 & 76.29 & \textbf{72.67}(\textcolor{red}{0.00}) & 59.09 & 62.62 & 56.03 & 59.24(\textcolor{red}{-0.28}) & \textbf{0.25}\\
\cline{1-15}
\multirow{4}{*}{ACoL} & Minmax & 55.42 & 64.45 & 61.20 & 60.36(\textcolor{red}{-0.60}) & 64.18 & 60.63 & 74.78 & 66.53(\textcolor{red}{-0.88}) & 53.93 & 57.04 & 57.66 & \textbf{56.21}(\textcolor{red}{0.00}) & 0.20\\
& Max & 55.25 & 64.44 & 61.17 & 60.28(\textcolor{red}{-0.67}) & 64.22 & 60.58 & 74.66 & 66.49(\textcolor{red}{-0.93}) & 53.91 & 57.06 & 57.63 & 56.20(\textcolor{red}{-0.01}) & 0.23\\
& PaS & 56.42 & 64.68 & 61.76 & \textbf{60.95}(\textcolor{red}{0.00}) & 63.83 & 60.30 & 74.74 & 66.29(\textcolor{red}{-1.13}) & 51.35 & 53.11 & 54.20 & 52.89(\textcolor{red}{-3.32}) & 2.85\\
& IVR & 55.43 & 65.08 & 61.20 & 60.57(\textcolor{red}{-0.39}) & 64.48 & 60.83 & 76.94 & \textbf{67.42}(\textcolor{red}{0.00}) & 53.62 & 56.89 & 57.07 & 55.86(\textcolor{red}{-0.35}) & \textbf{0.06}\\
\cline{1-15}
\multirow{4}{*}{SPG} & Minmax & 59.54 & 64.45 & 63.62 & 62.54(\textcolor{red}{-1.42}) & 67.74 & 64.17 & 71.98 & 67.96(\textcolor{red}{-2.18
}) & 58.98 & 63.68 & 58.33 & \textbf{60.33}(\textcolor{red}{0.00}) & 1.22\\
& Max & 59.22 & 64.35 & 63.40 & 62.32(\textcolor{red}{-1.64}) & 66.94 & 63.65 & 70.73 & 67.11(\textcolor{red}{-3.04}) & 58.64 & 63.64 & 58.51 & 60.26(\textcolor{red}{-0.07}) & 2.20\\
& PaS & 61.40 & 65.16 & 65.30 & \textbf{63.95}(\textcolor{red}{0.00}) & 64.46 & 61.07 & 69.39 & 64.97(\textcolor{red}{-5.17}) & 55.22 & 58.84 & 55.22 & 56.43(\textcolor{red}{-3.91}) & 7.26\\
& IVR & 60.22 & 65.38 & 64.09 & 63.23(\textcolor{red}{-0.72}) & 71.06 & 65.01 & 74.36 & \textbf{70.14}(\textcolor{red}{0.00}) & 58.81 & 63.59 & 57.72 & 60.04(\textcolor{red}{-0.29}) & \textbf{0.19}\\
\cline{1-15}
\multirow{4}{*}{ADL} & Minmax & 63.67 & 62.55 & 64.74 & 63.65(\textcolor{red}{-1.10}) & 69.40 & 66.66 & 69.03 & 68.36(\textcolor{red}{-0.27}) & 58.93 & 57.01 & 57.01 & \textbf{57.65}(\textcolor{red}{0.00}) & 0.33\\
& Max & 63.43 & 62.47 & 64.44 & 63.45(\textcolor{red}{-1.31}) & 69.28 & 66.45 & 69.06 & 68.26(\textcolor{red}{-0.37}) & 58.42 & 57.04 & 56.84 & 57.43(\textcolor{red}{-0.22}) & 0.35\\
& PaS & 64.52 & 63.58 & 66.17 & \textbf{64.76}(\textcolor{red}{0.00}) & 68.12 & 63.78 & 67.17 & 66.36(\textcolor{red}{-2.28}) & 55.48 & 53.86 & 54.21 & 54.52(\textcolor{red}{-3.13}) & 2.62\\
& IVR & 64.28 & 64.07 & 65.44 & 64.60(\textcolor{red}{-0.16}) & 69.50 & 66.94 & 69.46 & \textbf{68.63}(\textcolor{red}{0.00}) & 58.84 & 56.75 & 56.49 & 57.36(\textcolor{red}{-0.29}) & \textbf{0.01}\\
\cline{1-15}
\multirow{4}{*}{CutMix} & Minmax & 59.10 & 65.01 & 64.61 & 62.90(\textcolor{red}{-1.45}) & 78.22 & 63.68 & 71.17 & 71.02(\textcolor{red}{-1.01}) & 59.37 & 64.41 & 60.63 & \textbf{61.47}(\textcolor{red}{0.00}) & 0.56\\
& Max & 58.39 & 64.97 & 64.33 & 62.56(\textcolor{red}{-1.79}) & 77.95 & 63.48 & 70.55 & 70.66(\textcolor{red}{-1.38}) & 58.89 & 64.58 & 60.66 & 61.38(\textcolor{red}{-0.09}) & 0.78\\
& PaS & 61.18 & 65.80 & 66.10 & \textbf{64.36}(\textcolor{red}{0.00}) & 74.39 & 61.71 & 70.77 & 68.96(\textcolor{red}{-3.08}) & 55.78 & 59.73 & 55.85 & 57.12(\textcolor{red}{-4.35}) & 5.00\\
& IVR & 59.33 & 65.95 & 65.18 & 63.49(\textcolor{red}{-0.87}) & 78.93 & 64.78 & 72.40 & \textbf{72.04}(\textcolor{red}{0.00}) & 59.24 & 63.98 & 60.20 & 61.14(\textcolor{red}{-0.33}) & \textbf{0.27}\\
\Xhline{3\arrayrulewidth}
\end{tabular}
\label{tab:nwcperf}}
\end{table*}

\end{document}